\pdfoutput=1

\documentclass[11pt]{article}

\usepackage[]{naacl2021}

\usepackage{times}
\usepackage{latexsym}

\usepackage[T1]{fontenc}
\usepackage[utf8]{inputenc}

\usepackage{microtype}

\usepackage{amsmath}
\usepackage{booktabs}
\usepackage{multirow}
\usepackage{graphicx}

\title{Cross-lingual Entity Alignment with Adversarial Kernel Embedding and Adversarial Knowledge Translation}

\author{$\text{Gong Zhang}^1, \text{Yang Zhou}^2, \text{Sixing Wu}^3, \text{Zeru Zhang}^2, \text{Dejing Dou}^1$\\
  $^{1}\text{University of Oregon, US} $\\
  $^{2}\text{Auburn University, US}$  \\
  $^{3}\text{Peking University, China}$ \\
  \{gzhang7, dou\}@cs.uoregon.edu, \{yangzhou, zzz0054\}.auburn.edu, wusixing@pku.edu.cn
  }

\begin{document}
\maketitle
\begin{abstract}
Cross-lingual entity alignment, which aims to precisely connect the same entities in different monolingual knowledge bases (KBs) together, often suffers challenges from feature inconsistency to sequence context unawareness. This paper presents a dual adversarial learning framework for cross-lingual entity alignment, DAEA, with two original contributions. First, in order to address the structural and attribute feature inconsistency between entities in two knowledge graphs (KGs), an adversarial kernel embedding technique is proposed to extract graph-invariant information in an unsupervised manner, and project two KGs into the common embedding space. Second, in order to further improve successful rate of entity alignment, we propose to produce multiple random walks through each entity to be aligned and mask these entities in random walks. With the guidance of known aligned entities in the context of multiple random walks, an adversarial knowledge translation model is developed to fill and translate masked entities in pairwise random walks from two KGs. Extensive experiments performed on real-world datasets show that DAEA can well solve the feature inconsistency and sequence context unawareness issues and significantly outperforms thirteen state-of-the-art entity alignment methods.

% Cross-lingual entity alignment serves an important role in transferring and augmenting knowledge of linguistic sources. Existing embedding-based methods usually align equivalent entities by first modeling entities from knowledge graph (KG) structures, and then mapping them into a unified vector space in which alignment can be measured as their distances to each other. Oftentimes, the N-hop neighborhood of an entity and alignment seeds are utilized as resources to perform the first and second step respectively, but these methods generally suffer from neighborhood structural heterogeneity and alignment seeds overfitting issues. In this paper, we present a novel cross-lingual entity alignment framework with \textbf{C}onsensus Graph Embedding and \textbf{C}omprehensive \textbf{C}ontext (DAEA) to tackle those issues. 
% %DAEA includes three components to learn alignment-oriented entity representation. 
% DAEA first incorporates multi-aspect information regarding entities under a unified encoding process, including both structure and attribute information. It then adopts an adversarial module to reach consensus embeddings by aligning the cross-lingual representation spaces. It further enriches the embeddings with comprehensive context of entities, using a novel knowledge translation module operating on a pair of random walks as a minimax game. Extensive experiments show that our method outperforms most of state-of-the-art entity alignment methods on three real world datasets.
\end{abstract}

\section{Introduction}

\begin{figure}[ht]
  \centering
  \includegraphics[width=0.9\linewidth]{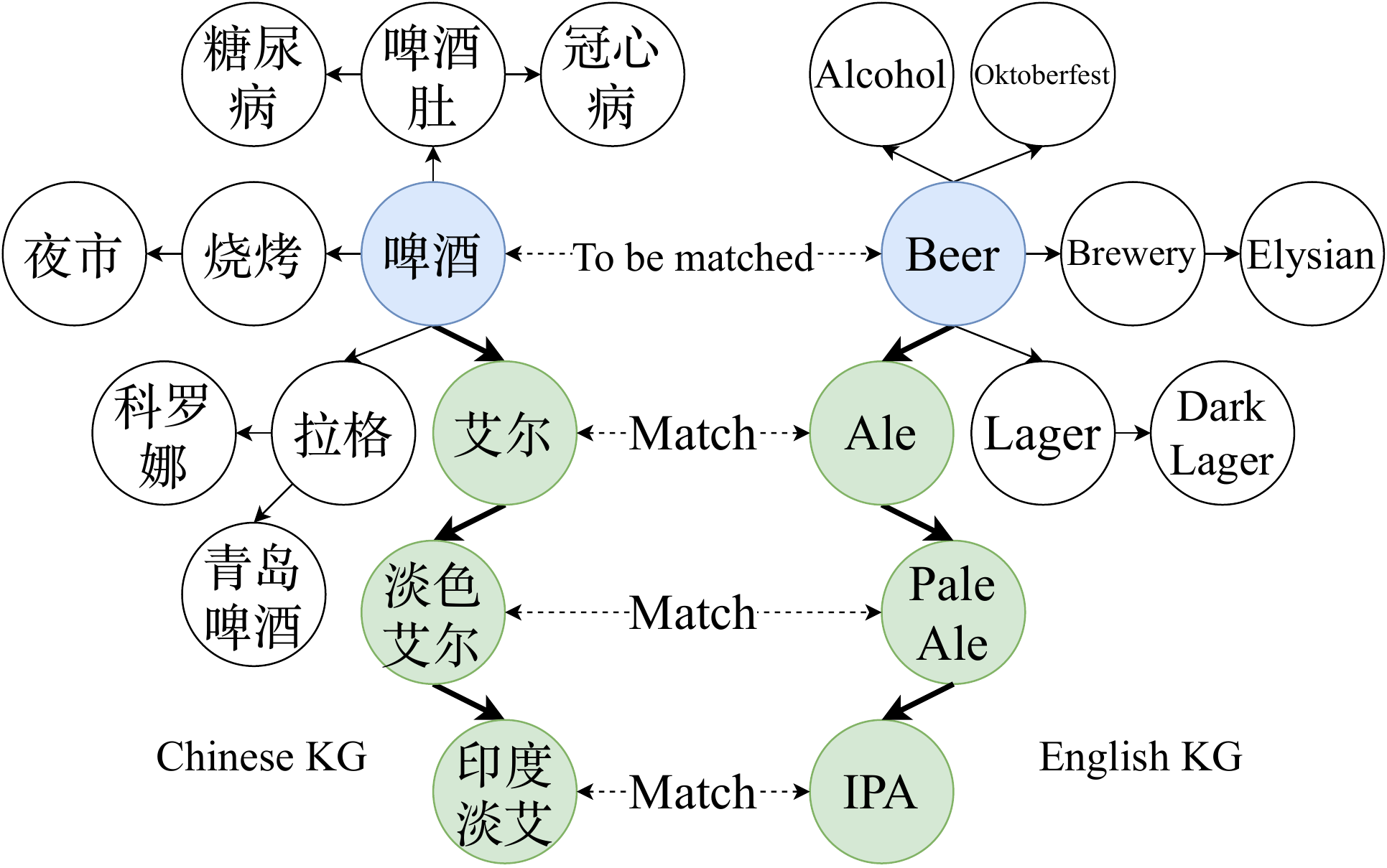}
  \caption{Illustrative examples: entity alignment in cross-lingual KGs with inconsistent neighbors and long-sequence context.}
\end{figure}

Cross-lingual entity alignment has become a powerful tool to automatically construct and complete multilingual knowledge bases, such as WordNet \cite{wordnet}, DBpedia \cite{auer2007dbpedia}, NELL \cite{nell}, BabelNet \cite{babelnet}, YAGO \cite{yago}, and ConceptNet \cite{conceptnet}. We have witnessed an impressive amount of work on cross-lingual entity alignment to revolutionize the understanding of the knowledge in a transformative manner \cite{transe, rdgcn}. Despite the remarkable performance of existing models, many multilingual knowledge bases are still far from complete. For example, the number of Japanese knowledge entities is only 2.5\% of the number of English knowledge entities in ConceptNet \cite{otani2018cross}. The coverage of cross-lingual knowledge in Wikipedia is less than 15\% \cite{wikipedia-coverage}.

Traditional cross-lingual entity alignment techniques are based on the fundamental assumption of topological consistency and/or attribute consistency \cite{consistency-assumption-1, consistency-assumption-2}: Two entities in different KGs are more likely to become an alignment if they share the similar topological and/or attribute features in respective KGs. However, it is often observed that the same entities in different KGs have diverse neighbors and attributes, due to asynchronous coverage of monolingual KBs and their heterogeneity \cite{nmn, heterogeneity}. Figure 1 provides a toy example of two KGs with inconsistent features, say {\it Beer} in English and Chinese associated with different neighbors. In this case, it is difficult to identify an alignment between them due to inconsistent structural features. Can we develop an unsupervised distribution matching model to project multiple KGs with inconsistent structural and attribute features into the common space for alleviating the feature inconsistency?

The majority of existing approaches utilize local neighborhood and attribute information to align a pair of entities at a time \cite{hman, cea}. When the languages in two KGs are quite different in coverage and grammar, the local information is insufficient for correctly aligning the entities, say the {\it Beer} example. However, we argue that long-sequence context may offer a great opportunity to improve the successful rate of entity alignment. In Figure 1, there are two colored paths with three known alignments of {\it Ale}, {\it Pale Ale}, {\it IPA} and their Chinese counterparts. If we utilize these three known alignments as guidance to match two paths, then it is highly possible to infer that English and Chinese {\it Beers} should be an alignment.

Different from early translation-based methods, which rely on 1-hop neighborhood information for entity alignment \cite{transe,iptranse, mtranse, bootea}, recent GCN-based models consider n-hop neighborhood information for improving the alignment quality \cite{gcn-align, hopgcn, nmn, alinet, mraea,entity-pair}. Other approaches introduce additional attribute information to enhance the alignment performance \cite{multike, COTSAE, BERT-INT}. However, how to effectively integrate the above multi-aspect information and long-sequence context into a unified model for better alignment is still an open question.

To address the above challenges, we develop a novel cross-lingual entity alignment method, DAEA, with two original contributions. First, we utilize graph convolutional networks (GCNs) to integrate the information of entities, relations, attributes, and entity name embeddings to learn a unified latent representation. An adversarial kernel embedding technique is proposed to project two KGs into a common Reproducing Kernel Hilbert Space (RKHS) induced with an adversarially trained kernel. The Maximum Mean Discrepancy (MMD) under a bounded function is employed to match the distributions of two KGs in the RKHS without known matched entities, such that two KGs are close to each other. Second, for each entity to be aligned, we produce multiple random walks to capture its 10-hop neighborhood information and long-sequence context. The random walks prefer to walk through known aligned entities, such that each random walk contains more aligned entities. We mask entities to be aligned in random walks and use unmasked entities (i.e., known aligned entities) as guidance to learn the entity alignment. An adversarial knowledge translation model is developed to fill and translate masked entities in pairwise random walks from two KGs.

Extensive evaluation on real datasets demonstrates the outstanding capability of DAEA to address the cross-lingual entity alignment problem against several state-of-the-art models.

\section{Our Approach}

A KG is defined as $G = (E, R, A)$, where $E, R, A$ are sets of entities, relations, and attributes. Given two heterogeneous KGs, $G_{source} = (E_s, R_s, A_s )$ and $G_{target} = (E_t, R_t, A_t )$ along with a set of known alignment seeds in form of entity pairs $\{(e_{si}, e_{ti}) | e_{si} \in E_s, e_{ti} \in E_t\}$. The entity alignment task is to find all equivalent entity pairs between source and target KGs.

\subsection{Overall Architecture}

\begin{figure*}[ht]
  \centering
  \includegraphics[width=\linewidth]{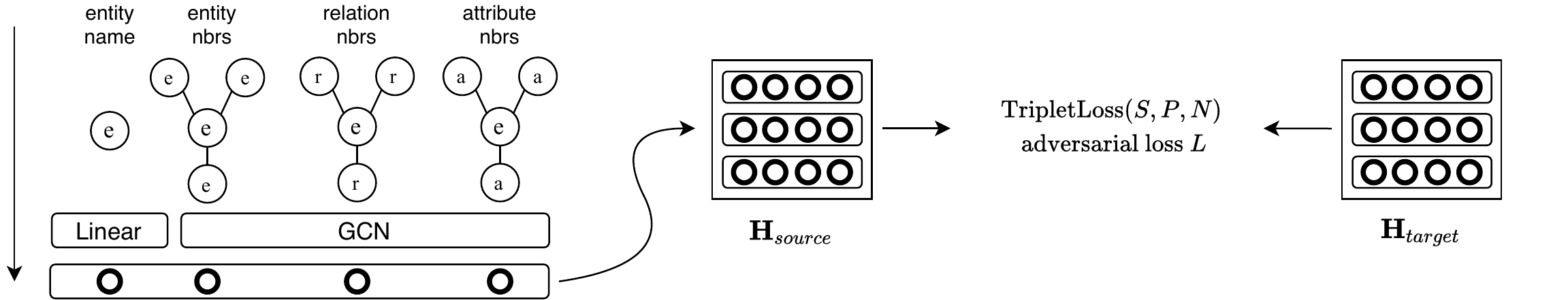}
  \caption{Integration of multi-aspect information and adversarial kernel embedding.}
\end{figure*}

DAEA aims to generate consensus and comprehensive context-aware embeddings for entity alignment, in which equivalent entities are close to each other. As depicted in Figure 2, it first takes 4 kinds of features as multi-aspect information input. All features are processed by GCNs under a unified entity-neighbors view, except for entity name. Both source and target KGs share the same GCN encoder to generate embeddings, which alleviates the disparity between their embedding distributions. Next, it adversarially matches the two embedding distributions with MMD as the distance metric. Then, it uses multiple random walks to enrich embeddings with comprehensive context as a knowledge translation process between source and target KGs; see Figure 3. 

Therefore, DAEA follows a 3-stage pipeline: (1) integration of multi-aspect information; (2) distribution matching; (3) knowledge translation. The first two stages are jointly trained and optimized to generate consensus embeddings, while stage 3 adopts the generated embeddings and is trained independently.

\subsection{Integration of Multi-Aspect Information}

To learn the multi-aspect information embeddings, DAEA takes 4 kinds of input features: entity name, structures, relations, and attributes. It utilizes GCNs to aggregate those features in a unified entity-neighbors view.

We use LASER\footnote{https://github.com/facebookresearch/LASER}, a sentence representation tool, to initialize the embeddings of those 4 features based on their literal labels. This strategy has been adopted by previous works \cite{nmn, hopgcn, rdgcn}, and it improves embeddings by introducing semantic information. For structures, relations and attributes features to work with GCNs, we uniformly model them in an entity-neighbors view. We define three adjacency matrices, entity adjacency matrix $\in \mathcal{R}^{N_e \times N_e}$, relation adjacency matrix $\in \mathcal{R}^{N_e \times N_r}$, and attribute adjacency matrix$\in \mathcal{R}^{N_e \times N_a}$, where $N_e$, $N_r$, $N_a$ are numbers of entities, relations, and attributes respectively. All of them are entity-centric, representing an entity through its entity neighbors, relation neighbors, and attribute neighbors. Then, we use 3 different GCN encoders to encode them respectively. The 3 GCN encoders are shared by source and target KGs for transferring information between them. Outputs of the GCNs plus entity name embeddings are fused together to form a single representation of an entity.  %We call these domain-invariant representations.
The fusion process can take various forms, such as mean, sum, and concatenation; we will discuss the effect of each one in Section 4.6. $\mathbf{H_{s}} \in \mathbf{R}^{N_s \times D}$ and $\mathbf{H_{t}} \in \mathbf{R}^{N_t \times D}$ are embedding matrices for source KG and target KG, where $D$ specifies the feature dimensions and $N$ is the number of entities. Each row in $\mathbf{H_{s}}$ (or $\mathbf{H_{t}}$) is the embedding of an entity $e$ in $G_{source}$ (or in $G_{target}$).

\begin{figure*}[h]
  \centering
  \includegraphics[width=0.85\textwidth]{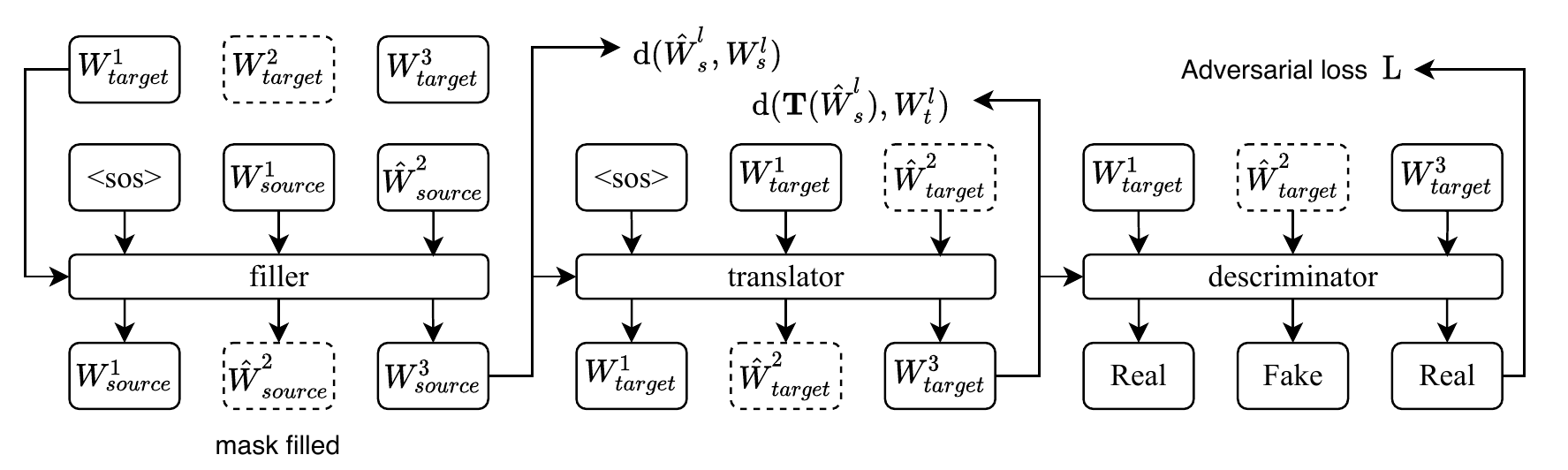}
  \caption{Adversarial knowledge translation. Dotted node represents a non-anchor position in a random walk. }
\end{figure*}

\subsection{Adversarial Kernel Embedding}

The Maximum Mean Discrepancy (MMD) metric is a linear kernel to measure the distance between distributions \cite{mmd1, mmd2}. The Generative adversarial network (GAN) technique was proposed to generate real-looking fake images \cite{gan,DBLP:conf/nips/SalimansGZCRCC16}. As a result, real and fake images tend to have the same distribution. This motivates us to propose to train the MMD kernel embedding of the distributions of cross-lingual KGs in an adversarial learning manner for matching their distributions in RKHS by continuously moving and twisting their distributions until the distributions finally overlap together, as well as to project the same concepts, relations, and knowledge into the common representation space for ease of translation among cross-lingual KGs.

We match the distributions of $G_{source}$ and $G_{target}$ by leveraging the kernel embedding technique to minimize the MMD between the latent representations $\mathbf{H_{s}}$ and $\mathbf{H_{t}}$, based on the MMD theory that an arbitrary data distribution can be uniquely denoted as an entry in a RKHS through the expectation map \cite{mmd1}. Specifically, an expectation map of $\mathbf{H_{s}}$ (or $\mathbf{H_{t}}$) is defined as $\mu[p] = \mathbf{E}_{v \sim p(v)} K(v,\cdot)$ (or $\mu[q] = \mathbf{E}_{u \sim p(u)} K(u,\cdot)$), where $f$ is a function defined in a universal RKHS $H$ with associated kernel $K(\cdot,\cdot)$. Then the MMD distance between $\mathbf{H_{s}}$ and $\mathbf{H_{t}}$ is calculated as follows.

{\small
\begin{equation} \label{eq:MMD}
\mathrm{MMD}(\mathbf{H_s},\mathbf{H_t}) = \mathrm{sup}_{f \in H} \langle \mu[p] - \mu[q], f \rangle = ||\mu[p] - \mu[q]||_H
\end{equation}
}

However, it is difficult to directly achieve the minimum of the MMD between $\mathbf{H_{s}}$ and $\mathbf{H_{t}}$. We propose an adversarial learning approach to approximately minimize the MMD for matching the distributions of $G_{source}$ and $G_{target}$. We introduce a neural network $F$ to simulate the kernel function $f$ in Eq.(\ref{eq:MMD}) and rewrite the MMD as follows. 

{\small
\begin{equation} \label{eq:MMD1}
\mathrm{MMD}(\mathbf{H_s},\mathbf{H_t})  = \\
  \max_{F} \Big|\Big|\frac{1} {N_s} F(\mathbf{H_s})^T \mathbf{1_s} - \frac{1} {N_t} F(\mathbf{H_t})^T \mathbf{1_t} \Big|\Big|_F^2
\end{equation}
}

\noindent where $\mathbf{1_s}$ and $\mathbf{1_t}$ are all-one vectors with the size of $N_s \times 1$ and $N_t \times 1$ respectively. $\frac{1} {N_s} F(\mathbf{H_s})^T \mathbf{1_s}$ and $\frac{1} {N_t} F(\mathbf{H_t})^T \mathbf{1_t}$ are the empirical measure of $\mu[p]$ and $\mu[q]$ respectively \cite{mmd1}.

The following GAN-based minimax game is designed to match the distributions of $G_{source}$ and $G_{target}$. In our model, $F$ is similar to a discriminator to maximize the MMD between $\mathbf{H_{s}}$ and $\mathbf{H_{t}}$, distinguishing if a distribution is from $G_{source}$ or $G_{target}$, while GCN encoders are compared to a generator to minimize MMD, so that they are able to generate indistinguishable representations of $\mathbf{H_{s}}$ and $\mathbf{H_{t}}$.

{\small
\begin{equation} \label{eq:GAN}
L = \min_{E} \max_{F} \Big|\Big|\frac{1} {N_s} F(\mathbf{H_t})^T \mathbf{1_s} - \frac{1} {N_t} F(\mathbf{H_t})^T \mathbf{1_t} \Big|\Big|_F^2
\end{equation}
}

The kernel function $F$ and the GCN encoders are trained together. We use alignment seeds and triplet loss with negative sampling as additional guides to supervise the optimization of GCN encoders. The loss of GCN encoders (generator) is: 

{\small
\begin{equation}
Loss = \mathrm{TripletLoss}(S, P, N) + L
\end{equation}
}

\noindent where $S$ are alignment seeds, $P$ are positive alignments from seeds, and $N$ are sampled negative alignments. Triplet loss \cite{tripletloss} measures relative similarity between positive and negative samples, optimizing the distances of alignments seed embeddings to be closer to each other rather than random negative samples.

\subsection{Adversarial Knowledge Translation}

In this section, we will use the entity embedding $\mathbf{H_{s}}$ and $\mathbf{H_{t}}$ to learn cross-lingual knowledge translation.
% which bridges the gap between distributions of embedding space.
It first randomly generates a random walk of length $L$ beginning from an anchor node $v_{t}$ in $G_{target}$, denoted by $W_{t}$. We use symbol $W^{l}_{t}$ to represent the $l^{th}$  $(1 \leq l \leq L)$ node in a random walk starting from $v_{t}$. 

Ideally, each sampled random walk would have balanced numbers of anchor nodes and non-anchor nodes. If a walk has only anchor nodes, knowledge translation will learn nothing more than what GCN encoder learns. On the other hand, if a walk contains only non-anchor nodes, the model will not have enough contextual information to make correctly decisions. To make anchor nodes and non-anchor nodes appear alternatively, if the current node is a non-anchor node, we set a probability of 0.9 to sample an adjacent anchor node, with only 0.1 probability to sample another non-anchor node and vice- versa. Namely, we don't force a walk to go between one anchor and one non-anchor alternatively, because if there is no new available anchor nodes(or non-anchor nodes), the walk may bounce back and forth within a pair of anchor and non-anchor nodes, making no progress in further exploring the KG. Therefore, the 10\% chance of breaking alternative sampling is meant to alleviate this situation.

For a random walk $W_{t} = (W^{1}_{t}, \cdots, W^{L}_{t})$ in $G_{target}$, a masked random walk $W_{s} = (W^{1}_{s}, \cdots, W^{L}_{s})$ in $G_{source}$ with a binary mask of the same length $M = (M_{1}, \cdots, M_{L})$ is produced where each $M_{l} \in \{0, 1\}$ is equal to 1 if $W^{l}_{t}$ is an anchor node or 0 otherwise. The node $W^{l}_{s}$ is then replaced with an empty identifier $\phi$ denoting a non-anchor node to be filled and translated if the corresponding mask is 0, and $W^{l}_{s}$ is substituted with the corresponding anchor node in $G_{source}$ if the mask is 1.

We use GAN to fill the masked nodes in a random walk $W_{s}$ and translate the entire $\hat{W}_{s}$ back into the original random walk $\hat{W}_{t}$. Specifically, a discriminator $\mathbf{D}(W^{i}_{t})$ outputs the probability of a node in random walk being a true node in $G_{target}$ rather than a fake node $\hat{W}^{l}_{t} = \mathbf{T}(\mathbf{F}(W^{l}_{s}))$ through the fill and translation of a masked walk $W_{s}$, where $\mathbf{F}$ is a filler to fill in the masks in $W^{i}_{s}$ with suitable nodes in $G_{source}$ and $\mathbf{T}$ is a translator to translate a filled walk $\hat{W}_{s} = \mathbf{F}(W_{s})$ into a walk $\hat{W}_{t} = \mathbf{T}(\hat{W}_{s})$. The combination of $\mathbf{T}$ and $\mathbf{F}$ forms the generator.

{\small
\begin{equation} \label{eq:MaskGAN}
\text{L} =  \min \limits_{\mathbf{F}, \mathbf{T}} \max \limits_{\mathbf{D}} -\Big[\log \mathbf{D}(W_t) + \log \Big(1-\mathbf{D}\big(\mathbf{T}(\mathbf{F}(W_s))\big)\Big)\Big] 
\end{equation}
}
where $\text{L}$ is the adversarial loss.

The filler $\mathbf{F}$ computes and decomposes the distribution over the random walk into the distribution of each non-anchor node in the random walk in the context of the mask $M_{i}$.

{\small
\begin{equation} \label{eq:Filler}
\begin{split}
\mathbf{F}(W_s) &= P\big(\hat{W}_s \big| W_s, M\big)\\ 
&= \prod_{l=1}^L P\big(\hat{W}^{l}_s \big| \hat{W}^{1}_s, \cdots, \hat{W}^{l-1}_s, W_s, M\big)
\end{split}
\end{equation}
}

\noindent where each distribution $P\big(\hat{W}^{l}_{s} \big| \hat{W}^{1}_{s}, \cdots, \hat{W}^{l-1}_{s}, W_{s}, M\big)$ is defined as an entity representation itself in $G_{source}$. This is different from defining it as softmax over all entity nodes in $G_{source}$. The benefit of using it as entity representation is a differentiable process between filler and translator. Otherwise, reinforcement learning techniques, such as policy gradient, have to be employed to make model optimization fully differentiable.

The discriminator $\mathbf{D}$ computes and decomposes the probability of a node in random walk $W^{l}_{t}$ or $\hat{W}^{l}_{t}$ being true in the context of the mask $M$. $\mathbf{D}$ has the same framework as $\mathbf{F}$, except that the output is a probability of a node being true.

{\small
\begin{equation} \label{eq:Discriminator0}
%\hspace{-0.7cm}
\begin{split}
    \mathbf{D}(W_t) &= \prod_{l=1}^L P\big(W^{l}_t = true \big| W_t, M\big), \\
    \mathbf{D}(\hat{W}_t) &= \prod_{l=1}^L P\big(\mathbf{T}(\mathbf{F}(W^{l}_s)) = true \big| W_s, M\big)
\end{split}
\end{equation}
}

As shown in Figure 3, we use three different LSTMs to implement $\mathbf{F}$, $\mathbf{D}$, and $\mathbf{T}$ respectively.

To make the training of this GAN-based component stable and fast- converging, besides generator and discriminator adversarial loss, we use two additional regularization terms in optimization objectives, one for filler and one for translator. The filler regularization term requires the filled nodes be similar to the real ones in $G_{source}$, namely $\hat{W}^{l}_s \approx  W^{l}_s$. In the case of non-anchor nodes, we use $\hat{W}^{l}_s \approx  \tilde{W}^{l}_s$, where $\tilde{W}^{l}_s$ is 
obtained by finding top-1 alignment using $\mathbf{H_{s}}$ and $\mathbf{H_{t}}$. The translator regularization term requires the translated node to be similar to real ones, namely $\mathbf{T}(\hat{W}^{l}_s) \approx  W^{l}_t$. The full generator loss is defined as:

{\small
\begin{equation}
Loss = \text{L} + \sum_{l=1}^L \Big[\mathrm{d}(\mathbf{F}(W^{l}_s), W^{l}_s) + \mathrm{d}(\mathbf{T}(\mathbf{F}(W^{l}_s)), W^{l}_t)\Big]
\end{equation}
}

\noindent where $\mathrm{d}(\cdot)$ is a function that measures distance between two embeddings.

\subsection{Model Inference}

With $\mathbf{H_{s}}$, $\mathbf{H_{t}}$ and trained translator $\mathbf{T}$, we are ready to make entity alignment prediction by first giving each entity embedding in $\mathbf{H_{s}}$ to $\mathbf{T}$, then using translated embeddings to find the best match in $\mathbf{H_{t}}$. In practice, this is non-trivial, because $\mathbf{T}$ is a sequence model. Feeding sampled random walks into $\mathbf{T}$ will generate multiple translated embeddings for each entity, since an entity may appear in multiple walks and may appear multiple times in a single walk. They are considered as holistic contextual information of the entity.

To integrate those contextual information, we first define the number of anchor nodes in a walks as the confidence score of that walk. The assumption is the more anchor nodes a walk has, the more likely it will provide richer context for generating good translations on non-anchor nodes. Based on the walk confidence score, we design three strategies to consolidate multiple translated embeddings: (1) Select the one with the highest confidence score. (2) Average all. (3) Average all with confidence score as weight. We will discuss their effects in Experiment section.

\section{Experiment}

\subsection{Dataset}
\begin{table}[h]
    \scriptsize
    \centering
    \setlength{\tabcolsep}{4pt}
    \begin{tabular}{lcccccl}
    \toprule
    \multicolumn{2}{c}{DBP15K} & \#Ent.  & \#Rel. & \#Tri. & \#Ent. mul. & \#Tri. mul.\\ 
    \midrule
    \multirow{2}{*}{ZH-EN} & ZH & 66,469  & 2,830  & 153,926  & \multirow{2}{*}{147\% }& \multirow{2}{*}{154\% } \\
    & EN & 98,125  & 2,317  & 237,674  &  & \\ 
    \multirow{2}{*}{JA-EN} & JA & 65,744  & 2,043  & 164,373 & \multirow{2}{*}{145\% } & \multirow{2}{*}{141\% } \\
    & EN & 95,680  & 2,096  & 233,319  &  & \\ 
    \multirow{2}{*}{FR-EN} & FR & 66,858  & 1,379  & 192,191 & \multirow{2}{*}{158\% } & \multirow{2}{*}{145\% } \\
    & EN & 105,889 & 2,209  & 278,590  &  & \\ 
    \bottomrule
    \end{tabular}
    \caption{Statistics of the DBP15K dataset.}
\end{table}

\begin{table*}[ht]
    \scriptsize
    \centering
    \begin{tabular}{lccccccccl}\toprule
    \multirow{2}{*}{Models} & 
    \multicolumn{3}{c}{$\mbox{DBP}_{\text{ZH-EN}}$} & \multicolumn{3}{c}{$\mbox{DBP}_{\text{JA-EN}}$} & \multicolumn{3}{c}{$\mbox{DBP}_{\text{FR-EN}}$}  \\\cmidrule(lr){2-4}\cmidrule(lr){5-7}\cmidrule(lr){8-10}
    & H@1 & H@10 & MRR & H@1 & H@10 & MRR & H@1 & H@10 & MRR \\ \midrule
    MTransE \cite{mtranse} & 0.308 & 0.614 & 0.364 & 0.279 &  0.575 & 0.349 & 0.244 & 0.556 & 0.335  \\
    IPTransE \cite{iptranse} & 0.406 & 0.735 & 0.516 & 0.367 & 0.693 & 0.474 & 0.333 & 0.685 & 0.451  \\
    Align-EA \cite{bootea} & 0.468 & 0.787 & 0.577 & 0.516 & 0.801 & 0.610 & 0.529 & 0.827 & 0.630 \\    
    BootEA \cite{bootea} & 0.608 & 0.846 & 0.691 & 0.576 & 0.830 & 0.663 & 0.610 & 0.853 & 0.694  \\ 
    RSN \cite{rsn} & 0.587 & 0.812 & 0.666 & 0.563 & 0.799 & 0.646 & 0.510 & 0.758 & 0.601  \\      
    \midrule
    GCN-Align \cite{gcn-align} & 0.421 & 0.745 & 0.533 & 0.424 & 0.761 & 0.541 & 0.414 & 0.778 & 0.537  \\   
    AliNet \cite{alinet} & 0.480 & 0.705 & 0.566 & 0.640 & 0.804 & 0.719 & 0.635 & 0.817 & 0.721 \\     
    HopGCN \cite{hopgcn} & 0.574 & 0.706 & 0.855 & 0.655 & 0.771 & 0.844 & 0.823 & 0.909 & 0.937  \\
    HMAN \cite{hman} & 0.572 & 0.828 & 0.673 & 0.552 & 0.860 & 0.660 & 0.488 & 0.821 & 0.604 \\     
    RDGCN \cite{rdgcn} & 0.698 & 0.842 & 0.751 & 0.762 & 0.894 & 0.810 & 0.879 & 0.956 & 0.907 \\      
    HGCN \cite{hgcn} & 0.718 & 0.860 & 0.767 & 0.751 & 0.894 & 0.801 & 0.888 & 0.961 & 0.915 \\      
    NMN \cite{nmn} & 0.733 & 0.869 & - & 0.785 & 0.912 & - & 0.902 & 0.967 & -  \\ 
    CEA \cite{cea} & 0.787 & - & - & 0.863 & - & - & \textbf{0.972} & - & -  \\    
    \midrule
    LASER & 0.622 & 0.709 & 0.583 & 0.719 & 0.802 & 0.705 & 0.825 & 0.881 & 0.817\\  
    MA & 0.784 & 0.893 & 0.793 & 0.858 & 0.938 & 0.847 & 0.930 & 0.974 & 0.945  \\     
    KE & 0.789 & 0.901 & 0.798 & 0.855 & 0.943 & 0.854 & 0.932 & \textbf{0.981} & 0.944  \\  
    DAEA & \textbf{0.828} & \textbf{0.924} & \textbf{0.843} & \textbf{0.870} & \textbf{0.951} & \textbf{0.882} & 0.936 & 0.971 & \textbf{0.948} \\    
    \bottomrule
    \end{tabular}
    \caption{Experiment Results: The upper part of the table names the baseline models, and the last entry in the table reports the performance of DAEA.}
\end{table*}

To evaluate the performance of DAEA on real-world datasets, we experiment it on DBP15K \cite{dbp15k}, a commonly used cross-lingual entity alignment task benchmark. It is built from 4 different language versions of DBpedia \cite{auer2007dbpedia}: English, Chinese, Japanese, and French. Three cross-lingual subset are provided: Chinese, Japanese, and French, each, to English. Every subset contains 15,000 aligned entity pairs. Follow the previous works \cite{hopgcn, bootea}, 30\% of aligned entity pairs are used for training, and the other 70\% are used for testing. A summary of DBP15K dataset statistics is shown in Table 1. It demonstrates that English, as by far the most well-resourced language, contains around 1.5 times the quantity of entities or triples present in Chinese, Japanese, and French, even though the latter three languages are generally not considered as low-resource languages. We adopt the popular Hits@topK and Mean Reciprocal Rank (MRR) as alignment evaluation metrics \cite{alinet}. Our source code and datasets are freely available online.\footnote{https://github.com/sunflower-the-cat/DAEA}

\subsection{Model variants}
To verify that each step of DAEA brings constructive effects towards entity alignment, we report performance at the following critical steps: (1) LASER embeddings of entity name, (2) Multi-aspect information embeddings (MA), (3) Adversarial kernel embedding (KE), (4) Adversarial knowledge translation (KT), where comprehensive context awareness is incorporated through random walks. Note that step 4 is the full model.

\subsection{Baselines}
To compare the overall performance of DAEA against other competitive alignment methods, we choose: MTransE \cite{mtranse}, IPTransE \cite{iptranse}, GCN-Align \cite{gcn-align}, BootEA \cite{bootea}), RSN \cite{rsn}, AliNet \cite{alinet}, RDGCN \cite{rdgcn}, HGCN \cite{hgcn}, HMAN \cite{hman}, HopGCN \cite{hopgcn}, Align-EA \cite{bootea}, NMN \cite{nmn}, CEA \cite{cea}. Among them, BootEA uses iterative-training procedure, in which newly aligned entities are used as training data for the next iteration.

\subsection{Experiment Results}

In table 1, we report the entity alignment performance of all comparable models on all DBP15K subsets. We divide the baseline methods into two categories: translation-based and GCN-based. It shows that the full implementation of DAEA outperforms all the baseline methods. The scores of DAEA are the average of multiple experiments. More specifically, our DAEA outperforms baseline models in terms of all subsets, except on DBP15K Fr-En. In comparison with the best baseline CEA, it increases the performance on the Zh-En subset by 0.04.

As one of the earliest attempts on entity alignment, MTransE sets the base Hit@1 score on the Zh-En subset at 0.308 by learning entity embeddings from triples. IPTransE improves ~0.1 over the based score by considering relational path information. BootEA delivers the best performance within the category, with the extension of iterative bootstrapping labeled alignments, which shows that the number of alignment seeds plays an important part in improving performance. RSN achieves the second-highest score, with only 0.02 lower than BootEA. It brings the long-term relational dependency information by using cross-graph random walks, which can capture more context information, compared with triple-only inputs.

The core part of GCN-based methods is incorporating n-hop neighborhood information. The first attempt of GCN-Align achieves 0.421, which already beats the performance of some translation-based methods. AliNet uses multi-hop neighborhood information to further boost the performance to 0.48 with only structural information. Starting from HopGCN, recent works not only refine the extraction of neighborhood information but also integrate additional information to improve alignment quality. HopGCN, RDGCN, HGCN and NMN all use pre-trained word embeddings to initialize GCN, while HMAN and CEA include attribute information and string similarity. The superior performance of these methods proves the significant effectiveness of integrating multiple sources of information.

\subsection{Ablation Study}

We focus on the ablation of 3 core modules in DAEA, along with entity name initialization. The 3 modules are MA (unify multi-aspect information), KE (mapping KG embeddings into an common space), and KT (introduce contextual awareness through random walks and translate between languages). Each of them is tested on top of previous modules to show their effectiveness.

\textbf{Entity Name Initialization.} We treat each entity name as a sentence and feed it into LASER, a Bidirectional LSTM model, to generate contextual embedding. It is worth noting that the Hits@1 scores of the entity names are comparable with some baseline models using word embedding. This shows the importance of label information as a feature, and the superiority of contextual sentence embedding over word embedding as an initialization method.

\textbf{MA.} It already outperforms all baselines in all subsets by nearly 0.03, except CEA. This performance leap shows that fusion of various features is the key when KGs have severe structural heterogeneity, e.g., imbalanced neighborhoods of a paired entity in different languages. Its improvements over HMAN also strengthen our argument that a unified procedure to incorporate features will benefit performance. 

\textbf{KE.} It aims to match embedding space distributions in an unsupervised manner using MMD. Though the performance increase is minor, the decrease of the distance between embedding spaces will help KT easily translate an entity from one language to another.

% \textbf{KT only} KT only MA \& KE directly uses entity name embeddings as input. We see a significant drop in performance. Like the results of RSN, this again proves better aligned cross-KG embeddings with abundant local information are crucial and complementary to global information implicitly residing in random walks. Together they achieve competitive performance.

\begin{table}[ht]
    \scriptsize
    \centering
    \setlength{\tabcolsep}{2pt}
    \begin{tabular}{lccccccccl}\toprule
    \multirow{2}{*}{Models} &  \multicolumn{3}{c}{$\mbox{DBP}_{\text{EN-ZH}}$} & \multicolumn{3}{c}{$\mbox{DBP}_{\text{EN-JA}}$} & \multicolumn{3}{c}{$\mbox{DBP}_{\text{EN-FR}}$} \\\cmidrule(lr){2-4}\cmidrule(lr){5-7}\cmidrule(lr){8-10}
    & H@1 & H@10 & MRR & H@1 & H@10 & MRR & H@1 & H@10 & MRR \\ 
    \midrule
    LASER  & 0.619 & 0.716 & 0.647 & 0.712 & 0.723 & 0.738 & 0.826 & 0.883 & 0.835 \\  
    MA & 0.769 & 0.875 & 0.803 & 0.841 & 0.923 & 0.873 & 0.934 & 0.971 & 0.952 \\  
    KE & 0.772 & 0.888 & 0.795 & 0.845 & 0.934 & 0.841 & 0.939 & 0.976 & 0.944 \\     
    DAEA & \textbf{0.888} & \textbf{0.961} & \textbf{0.913} & \textbf{0.910} & \textbf{0.970} & \textbf{0.924} & \textbf{0.952} & \textbf{0.980} & \textbf{0.969} \\    
    \bottomrule
    \end{tabular}
    \caption{DAEA variants on reversed DBP15K}
\end{table}

\textbf{KT.} Our full model further improves the Hits@1 performance on the Zh-En and Ja-En subsets. This resonates with our argument that using random walks to introduce contextual information strengthens the expressiveness of entity representation. In Table 3, we test the reversed alignment, e.g. En-Zh. It consistently gives better performance than their forward versions, e.g., Zh-En. Namely, entity alignment can be better performed by translating high- resource language entities to low- resource language entities. 
% We argue that the translator modifies the positions of the source entity in embedding space by pushing a source entity towards its counterpart target entity; and intuitively, it is easier to make alignment with fewer entities in the target domain.

\subsection{Analysis and Discussion}

\begin{figure}[h]
  \centering
  \includegraphics[width=\linewidth]{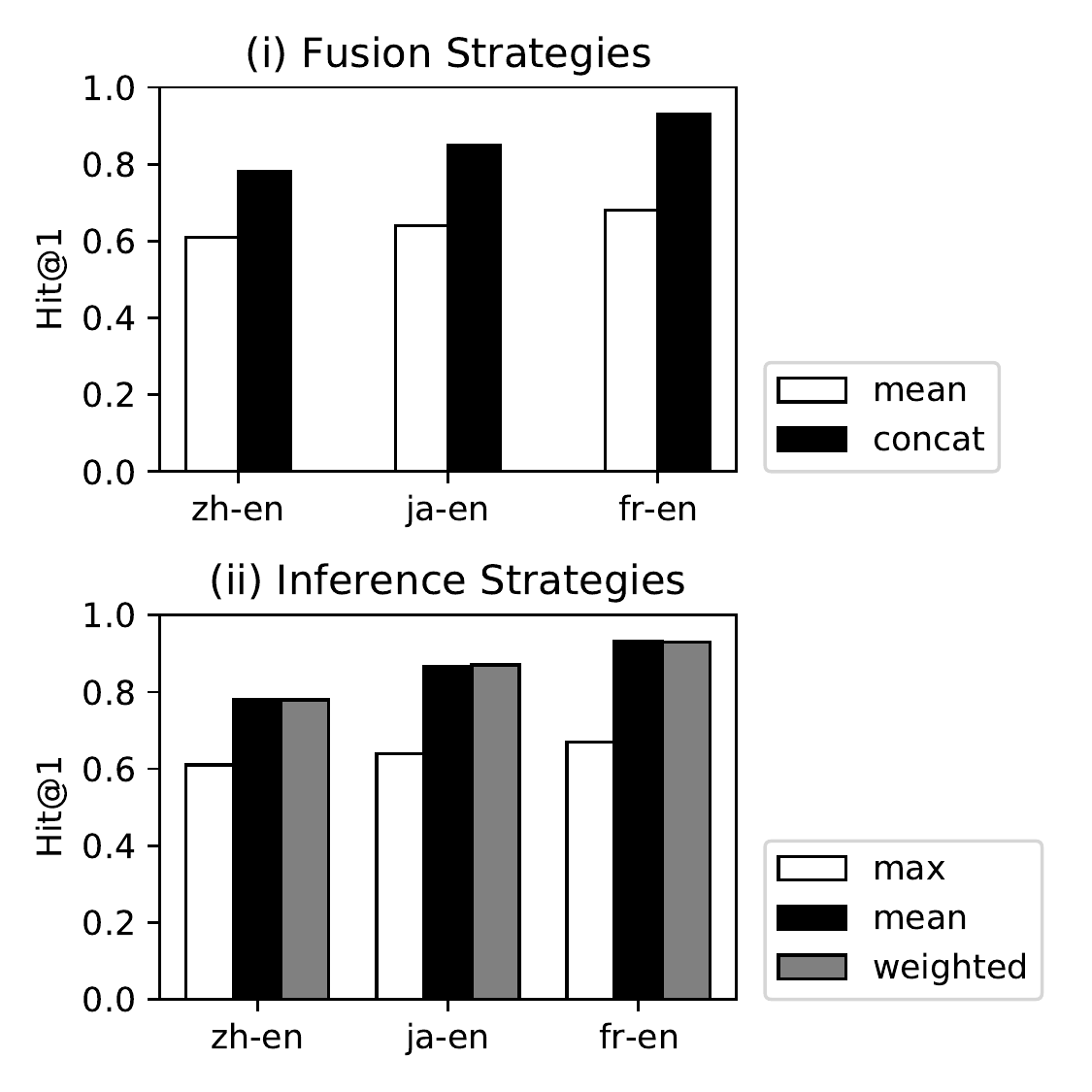}
  \caption{Fusion and inference strategies }
\end{figure}

\textbf{Fusion of GCN Encoders Outputs.} In the GCN encoder phase, we obtain three local structural representations of an entity. We test two fusion strategies on them: mean and concatenation. In Figure 4(i), results show that concatenation gives better performance than mean. This suggests that heterogeneity not only happens across different KGs, but also exists across different features within a single KG. Therefore, simply taking average to fuse embeddings of different features makes the fused representation unpredictable and hurting the overall model performance.

\textbf{Model Inference with Confidence Score.} We propose three strategies to integrate contextual information based on confidence score of random walk, as follows. (1) Selecting the representation in the walk with the highest confidence score. This essentially generate random result, because confidence score is defined per walk, not per entity, and the same entity may appear in a walk multiple times. It ends up picking random entity representation from the walk, giving inconsistent results. (2) Weighted average using confidence score. Each entity representation will receive, as its weight, the confidence score of the walk from which it comes. Then, all representations are averaged based on those weights after softmax. (3) Simple average. As indicated by Figure 4(ii), it turns out (2) and (3) produce consistent results with good performance. This observation shows that the same entity's representations, generated by the translator from various random walks, may emphasize different contextual information, and the aggregation of all those different emphases forms a holistic picture of an entity, thus giving good entity alignment results.

\subsection{Knowledge Completion}

\begin{table}[h]
    \scriptsize
    \centering
    \setlength{\tabcolsep}{2pt}
    \begin{tabular}{lccccccccl}
    \toprule
    \multirow{2}{*}{Models} & 
    \multicolumn{3}{c}{$\text{DBP}_{\text{ZH-EN}}$} & \multicolumn{3}{c}{$\text{DBP}_{\text{JA-EN}}$} & \multicolumn{3}{c}{$\text{DBP}_{\text{FR-EN}}$} \\\cmidrule(lr){2-4}\cmidrule(lr){5-7}\cmidrule(lr){8-10}
    & H@1 & H@10 & MRR & H@1 & H@10 & MRR & H@1 & H@10 & MRR \\
    \midrule
    AlignEA & 0.124 & 0.320 & 0.000 & 0.000 & 0.000 & 0.000 & 0.106 & 0.322 & 0.187 \\
    BootEA & 0.172 & 0.498 & 0.293 & 0.128 & 0.458 & 0.250 & 0.146 & 0.477 & 0.266 \\    
    GCN-Align & 0.108 & 0.283 & 0.177 & 0.076 & 0.279 & 0.154 & 0.091 & 0.272 & 0.159 \\    
    RDGCN & 0.168 & 0.492 & 0.287 & 0.145 & 0.499 & 0.275 & 0.182 & 0.596 & 0.330 \\     
    HMAN & 0.140 & 0.384 & 0.232 & 0.106 & 0.379 & 0.208 & 0.115 & 0.342 & 0.200 \\     
    HGCN & 0.174 & 0.504 & 0.295 & 0.136 & 0.491 & 0.268 & 0.139 & 0.541 & 0.281 \\     
    AliNet & 0.152 & 0.428 & 0.255 & 0.141 & 0.434 & 0.245 & 0.139 & 0.439 & 0.247 \\  
    \midrule
    DAEA & 0.189 & 0.525 & 0.314 & 0.170 & 0.587 & 0.322 & 0.197 & 0.661 & 0.362 \\
    \bottomrule
    \end{tabular}
    \caption{Knowledge Completion Results}
\end{table}

We further test the quality of alignments produced by DAEA on the knowledge completion task, which aims to reconcile the structural
differences by completing the missing relations \cite{mugnn}. It is performed by inputting generated entity alignments into the link prediction module of TransE \cite{transe}, which is based on sub-graphs built from correctly aligned entities. The results are reported in Table 4. DAEA consistently achieves the best score across all subsets.

\subsection{Case Study}

\begin{figure}[ht]
  \centering
  \includegraphics[width=0.8\linewidth]{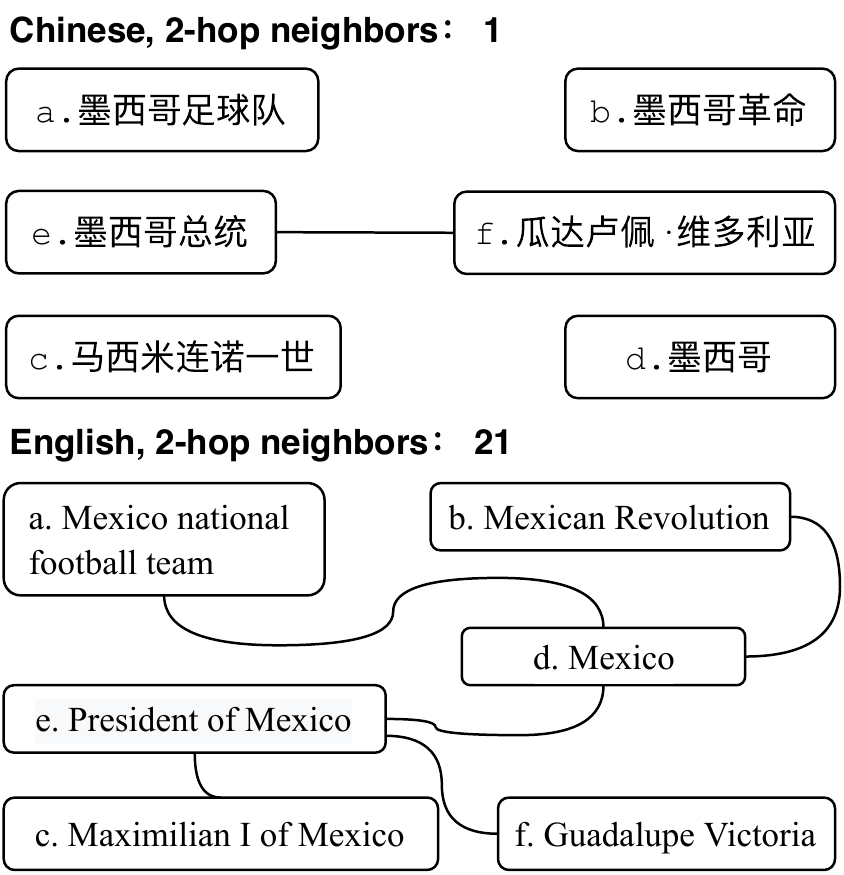}
  \caption{A case study of an entity pair with imbalanced neighborhoods.}
\end{figure}

Figure 5 shows an example, in DBP15K subset Zh-En, where a entity pair is indicated by its numbering. The entity pair (with shadow) of \textit{President of Mexico} should be aligned. Though related entities such as \textit{Mexico}, \textit{Mexico national football team} and \textit{Maximilian I of Mexico} appear in both KGs, they are not in the neighborhood of \textit{President of Mexico} in Chinese. The imbalance of neighborhood information in different languages make it hard to associate them with structural information only. DAEA utilizes multi-aspect information and contextual information from random walk pairs sampled from a high-resource language.  The contextual information that doesn't exist in a low-resource language. It finally translates the context to the low-resource language. Thus, our DAEA is able to correctly align them.

% $e_{zh}$ is one of many long tail entities with only one neighbor at each hop of first 2-hops. By contrast, $e_{en}$ has more neighbors, because English is a high-resource language.  For this example, though $e_{zh}$ to $e_{en}$ don't share much similarity within their immediate vicinities, they are all from the same broader topic, which shares much more similarities. Utilizing this comprehensive context information, our DAEA is able to correctly align them.

\section{Conclusion}

In this paper, we present a novel cross-lingual entity alignment framework, DAEA. It tackles feature inconsistency and sequence context unawareness issues by dual adversarial learning: (1) unsupervised adversarial kernel embedding to project KG features into common embedding space; and (2) adversarial knowledge translation on random walk pairs to incorporate sequence context awareness. Extensive experiment results on real-world datasets demonstrate that DAEA achieves competitive performance across languages and evaluation metrics.

\section*{Broader Impact and Ethics}

Real-world things are ubiquitously modeled as entities and relations in graph data, such as social networks, communication networks, scene graphs, and knowledge graphs. They are independently designed and developed for different domains, applications, and languages, while sharing common entities, yet in different representations. It is hugely beneficial to integrate them into a universal graph, which provides comprehensive information and connections of a unique, real- world entity in one place.

Entity alignment is the research topic aimed at finding common entities across heterogeneous graphs. It has been widely applied to many real-world scenarios, ranging from universal product recommendation \cite{recommendation-system-1, recommendation-system-2} and account linking in different social networks \cite{account-linking-1, account-linking-2}, to online shopping \cite{online-shopping}. Due to its great value in applications, more work is needed to improve alignment accuracy.

Among the potential risks of applying entity alignment techniques are privacy issues, especially for the task of account linking across social networks, in which certain user identity data are important toward performing alignment. Recent research papers on differential privacy and privacy-preserving graph analytics have shown promising results toward privacy protection. The combination of these techniques and entity alignment could offer an opportunity to generate high-quality alignment while protecting sensitive information about individuals.

\bibliography{anthology,custom}
\bibliographystyle{acl_natbib}

\end{document}